\documentclass[11pt,a4paper]{article}
\usepackage[hyperref]{acl2019}
\usepackage{times}
\usepackage{latexsym}
\usepackage[utf8]{inputenc} 
\usepackage[T1]{fontenc}    
\usepackage{hyperref}       
\usepackage{url}            
\usepackage{booktabs}       
\usepackage{amsfonts}       
\usepackage{graphicx}       
\usepackage{multicol}
\usepackage{url}

\newcommand\fire{\raisebox{-.4ex}{\protect\includegraphics[height=2.5ex]{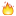}}}
\newcommand\frozen{\raisebox{-.4ex}{\protect\includegraphics[height=2.5ex]{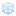}}}
\newcommand\blend{\raisebox{-.4ex}{\protect\includegraphics[height=2.5ex]{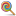}}}

\aclfinalcopy 

\title{Resolving Gendered Ambiguous Pronouns with BERT}

\author{Matei Ionita \\
  University of Pennsylvania  \\
  \texttt{matei@sas.upenn.edu} \\\And
  Yury Kashnitsky \\
  Moscow Institute of Physics and Technology \\
  \texttt{yury.kashnitsky@phystech.edu} \\\AND
  Ken Krige \\
  Kitsong \\
  \texttt{kenkrige@gmail.com} \\\And
  Vladimir Larin \\
  PJSC Sberbank \\
  \texttt{vlarine@gmail.com} \\\AND
  Denis Logvinenko \\
  Jet Infosystems \\
  \texttt{dennis\_l@rambler.ru} \\\And
  Atanas Atanasov \\
  Sofia University \\
  \texttt{a.atanasov1@ocado.com} }

\begin{document}
\maketitle

\begin{abstract}

Pronoun resolution is part of coreference resolution, the task of pairing an expression to its referring entity. This is an important task for natural language understanding and a necessary component of machine translation systems, chat bots and assistants. Neural machine learning systems perform far from ideally in this task, reaching as low as 73\% F1 scores on modern benchmark datasets. Moreover, they tend to perform better for masculine pronouns than for feminine ones. Thus, the problem is both challenging and important for NLP researchers and practitioners. In this project, we describe our BERT-based approach to solving the problem of gender-balanced pronoun resolution. We are able to reach 92\% F1 score and a much lower gender bias on the benchmark dataset shared by Google AI Language team.

\end{abstract}


\section{Introduction}
\label{sec:intro}

In this work, we are dealing with gender bias in pronoun resolution. A more general task of coreference resolution is reviewed in Sec. \ref{sec:coref_algos}. In Sec. \ref{sec:kaggle}, we give an overview of a related Kaggle competition. Then, Sec. \ref{sec:gap_paper} describes the GAP dataset and Google AI's heuristics to resolve pronomial coreference in a gender-agnostic way, so that pronoun resolution is done equally well in cases of masculine and feminine pronouns. In Sec. \ref{sec:our_solution}, we provide the details of our BERT-based solution while in Sec. \ref{sec:results} we analyze pleasantly low gender bias specific for our system (our code is shared on GitHub\footnote{\url{https://github.com/Yorko/gender-unbiased_BERT-based_pronoun_resolution}}). Lastly, in Sec. \ref{sec:future}, we draw conclusions and express some ideas for further research. 

\section{Related work}
\label{sec:coref_algos}

Among popular approaches to coreference resolution are:\footnote{\url{https://bit.ly/2JbKxv1}} rule-based, mention pair, mention ranking, and clustering. As for rule-based approaches, they describe naïve Hobbs algorithm \cite{Hobbs1986} which, in spite of being naïve, has shown state-of-the-art performance on the OntoNotes dataset\footnote{\url{https://catalog.ldc.upenn.edu/LDC2013T19}} up to 2010. 

Recent state-of-the-art approaches \cite{lee2018,e2e-coref,peters2018contextualized} are pretty complex examples of mention ranking systems. The 2017 version is the first end-to-end coreference resolution model that didn’t utilize  syntactic parsers or hand-engineered mention detectors. Instead, it used LSTMs and attention mechanism to improve over previous NN-based solutions.

%
%
%
%

Some more state-of-the-art coreference resolution systems are reviewed in \cite{gap_paper} as well as popular datasets with ambiguous pronouns: Winograd schemas \cite{Winograd}, WikiCoref \cite{WikiCoref}, and The Definite Pronoun Resolution Dataset \cite{dpr_dataset}. We also refer to the GAP paper for a brief review of gender bias in machine learning.

We further outline that e2e-coref model \cite{lee2018}, in spite of being state-of-the-art in coreference resolution, didn't show good results in the pronoun resolution task that we tackled, so we only used e2e-coref predictions as an additional feature.

\section{Kaggle competition ``Gendered Pronoun Resolution''}
\label{sec:kaggle}

Following Kaggle competition ``Gendered Pronoun Resolution'',\footnote{\url{https://www.kaggle.com/c/gendered-pronoun-resolution}} for each abstract from Wikipedia pages we are given a pronoun, and we try to predict the right coreference for it, i.e. to which named entity (A or B) it refers. Let’s take a look at this simple example: 

\textit{``John entered the room and saw [A] Julia. [Pronoun] She was talking to [B] Mary Hendriks and looked so extremely gorgeous that John was stunned and couldn’t say a word.''}

Here ``Julia'' is marked as entity A, ``Mary Hendriks'' – as entity B, and pronoun ``She'' is marked as Pronoun. In this particular case the task is to correctly identify to which entity the given pronoun refers. 

\begin{figure*}[!ht]
  \centering
  \includegraphics[width=\linewidth]{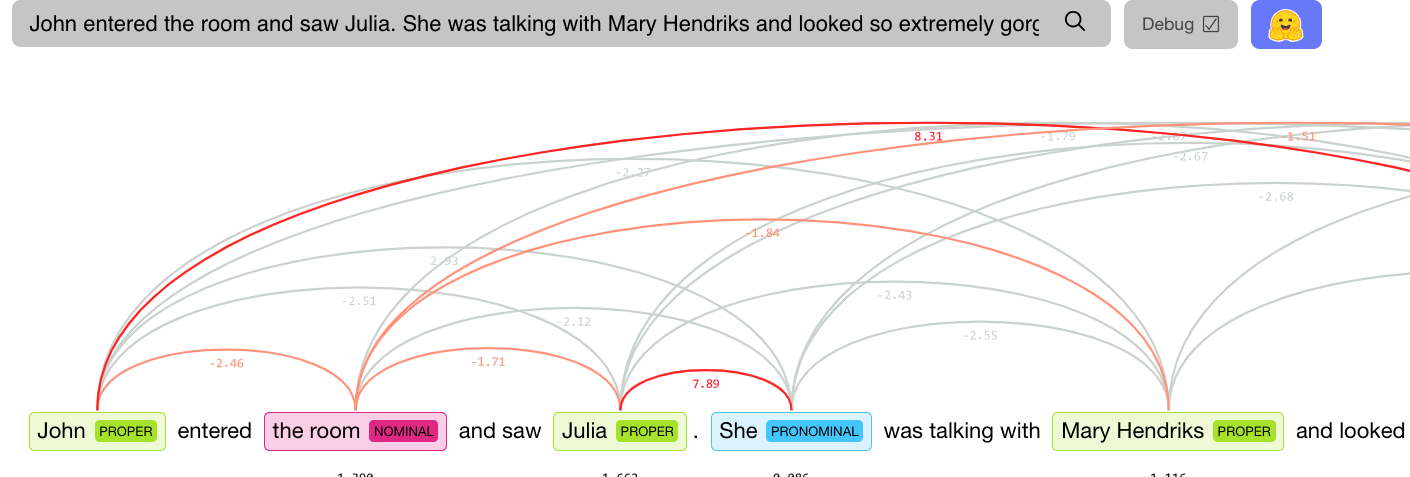}
  \caption{Coreference resolution visualized with HuggingFace demo \url{https://huggingface.co/coref/}.}
  \label{fig:huggingface_coref_demo2}
\end{figure*}

If we feed this sentence into a coreference resolution system (see Fig. \ref{fig:huggingface_coref_demo2} and online demo\footnote{\url{https://bit.ly/2I4tECI}}), we see that it correctly identifies that “she” refers to Julia, it also correctly clusters together two mentions of “John” and detects that Mary Hendriks is a two-word span.

For instance, if you take an abstract like this it’s pretty hard to resolve coreference.

\textit{``Roxanne, a poet who now lives in France. Isabel believes that she is there to help Roxanne during her pregnancy with her toddler infant, but later realizes that her father and step-mother sent her there so that Roxanne would help the shiftless Isabel gain some direction in life. Shortly after she (pronoun) arrives, Roxanne confides in Isabel that her French husband, Claude-Henri has left her.''}

Google AI and Kaggle (organizers of this competition) provided the GAP dataset \cite{gap_paper} with 4454 snippets from Wikipedia articles, in each of them named entities A and B are labeled along with a pronoun. The dataset is labeled, i.e. for each sentence a correct coreference is specified, one of three mutually-exclusive classes: either A or B or “Neither”. Thus, the prediction task is actually that of multiclass classification type. 

Moreover, the dataset is balanced w.r.t. masculine and feminine pronouns. Thus, the competition was supposed to address the problem of building a coreference resolution system which is not susceptible to gender bias, i.e. works equally well for masculine and feminine pronouns. 

These are the columns provided in the dataset \cite{gap_paper}:

\begin{itemize}
\item ID - Unique identifier for an example (matches to Id in output file format)
\item 	Text - Text containing the ambiguous pronoun and two candidate names (about a paragraph in length)
\item 	Pronoun - target pronoun (text)
\item 	Pronoun-offset - character offset of Pronoun in Text
\item 	A - first name candidate (text)
\item 	A-offset - character offset of name A in Text
\item 	B - second name candidate
\item 	B-offset - character offset of name B in Text
\item 	URL - URL of the source Wikipedia page for the example
\end{itemize}

Evaluation metric chosen for the competition\footnote{\url{https://www.kaggle.com/c/gendered-pronoun-resolution/overview/evaluation}} is multiclass logarithmic loss.  Each pronoun has been labeled with whether it refers to A, B, or ``Neither''. For each pronoun, a set of predicted probabilities (one for each class) is submitted. The formula is then

$$logloss = - \frac{1}{N}\sum_{i=1}^N\sum_{j=1}^M y_{ij}\log p_{ij},$$

\noindent where $N$ is the number of samples in the test set, $M$ is $3$,  log is the natural logarithm, $y_{ij}$ is $1$ if observation $i$ belongs to class $j$ and $0$ otherwise, and $p_{ij}$ is the predicted probability that observation $i$ belongs to class $j$.

Unfortunately, the chosen evaluation metric does not reflect the mentioned above goal of building a gender-unbiased coreference resolution algorithm, i.e. the metric does not account for gender imbalance - logarithmic loss may not reflect the fact that e.g. predicted pronoun coreference is much worse for masculine pronouns than for feminine ones. Therefore, we explore gender bias separately in Sec. \ref{sec:results} and compare our results with those published by the Google AI Language team (reviewed in Sec. \ref{sec:gap_paper}).

\section{Mind the GAP: A Balanced Corpus of Gendered Ambiguous Pronouns}
\label{sec:gap_paper}

Google AI Language team addresses the problem of gender bias in pronoun resolution (when systems favor masculine entities) and a gender-balanced labeled corpus of 8,908 ambiguous pronoun-name pairs sampled to provide diverse coverage of challenges posed by real-world text \cite{gap_paper} (further referred to as the GAP dataset). 
They run 4 state-of-the-art coreference resolution models \cite{Lee2013,clark-manning-2015-entity,wiseman-etal-2016-learning,e2e-coref} on the OntoNotes and GAP datasets reporting F1 scores separately for masculine and feminine pronoun-named entity pairs (metrics \textbf{M} and \textbf{F} in the paper). Also they measure ``gender bias'' defined as  \textbf{B = F / M}. In  general, they conclude, these models perform better for masculine pronoun-named entity pairs, but still pronoun resolution is challenging - all achieved F1 scores are less than $0.7$ for both datasets.

Further, they propose simple heuristics (called surface, structural and Wikipedia \textit{cues}). The best reported cues are ``Parallelism'' (if the pronoun is a subject or direct object, select the closest candidate with the same grammatical argument) and ``URL'' (select the syntactically closest candidate which has a token overlap with the page
title). They compare the performance of ``Parallelism + URL'' cue with e2e-coref \cite{e2e-coref} on the GAP dataset and, surprisingly enough, conclude that heuristics work better achieving better F1 scores ($0.742$ for \textbf{M} and $0.716$ for \textbf{F}) at the same time being less gender-biased (some of heuristics are totally gender-unbiased, for ``Parallelism + URL'' \textbf{B} = F / M $= 0.96$).

Finally, they explored Transformer architecture \cite{Vaswani2017} for this task and observed that the coreference signal is localized on specific heads and that these heads are in the deep layers of the network.  In Sec. \ref{sec:our_solution} we confirm this observation. Actually, they select the candidate which attends most to the pronoun (``Transformer heuristic'' in the paper). Even though they conclude that Transformer models implicitly learn language understanding relevant to coreference resolution, as for F1 scores, they didn't make it work better than e2e-coref or Parallelism cues (F1 scores lower that $0.63$). More to that, proposed Transformers heuristics are a bit biased towards masculine pronouns with \textbf{B} from $0.95$ to $0.98$.

Further we report a much stronger gender-unbiased BERT-based \cite{bert} pronoun resolution system.

\section{System}
\label{sec:our_solution}

BERT \cite{bert} is a transformer architecture, pre-trained on a large corpus (Wikipedia + BookCorpus), with 12 to 24 transformer layers. Each layer learns a 1024-dimensional representation of the input token, with layer 1 being similar to a standard word embedding, layer 24 specialized for the task of predicting missing words from context. At the same time BERT embeddings are learned for a second auxiliary task of resolving whether two consequent sentences are connected to each other or not. 

In general, motivated by \cite{tenney2019}, we found that BERT  provides very good token embeddings for the task in hand. 

Our proposed pipeline is built upon solutions by teams ``Ken Krige'' and  ``[ods.ai] five zeros'' (placed 5 and 22 in the final leaderboard\footnote{\url{https://www.kaggle.com/c/gendered-pronoun-resolution/leaderboard}} correspondingly). The way these two teams approached the competition task are described in two Kaggle posts.\footnote{\url{https://www.kaggle.com/c/gendered-pronoun-resolution/discussion/90668}}\footnote{\url{https://www.kaggle.com/c/gendered-pronoun-resolution/discussion/90431}} The combined pipeline includes several subroutines:

\begin{itemize}
\item Extracting BERT-embeddings for named entities A, B, and pronouns
\item Fine-tuning BERT classifier
\item Hand-crafted features
\item Neural network architectures
\item Correcting mislabeled instances
\end{itemize}

\subsection{Extracting BERT-embeddings for named entities A, B, and pronouns}
\label{subsec:bert_preprocess}
We concatenated embeddings for entities A, B, and Pronoun taken from Cased and Uncased large BERT ``frozen'' (not fine-tuned) models.\footnote{\url{https://github.com/google-research/bert}}
We noticed that extracting embeddings from intermediate layers (from -4 to -6) worked best for the task. Also we added pointwise products of embeddings for Pronoun and entity A, Pronoun and entity B as well as AB - PP. First of these embedding vectors expresses similarity between pronoun and A, the second one expresses similarity between pronoun and B, the third vector is supposed to represent the extent to which entities A and B are similar to each other but differ from the Pronoun. 

\subsection{Fine-tuning BERT classifier}
\label{subsec:finetune}
Apart from extracting embeddings from original BERT models, we also fine-tuned BERT classifier for the task in hand. We made appropriate changes to the ``run\_classifier.py'' script from Google's repository.\footnote{\url{https://github.com/google-research/bert}} Preprocessing input data for the BERT input layer included  stripping text to 64 symbols, then into 4 segments, running BERT Wordpiece for each segment, adding start and end tokens (with truncation if needed) and concatenating segments back together. The whole preprocessing is reproduced in a Kaggle Kernel\footnote{\url{https://www.kaggle.com/kenkrige/bert-example-prep}} as well as in our final code on GitHub.\footnote{\url{https://github.com/Yorko/gender-unbiased_BERT-based_pronoun_resolution}} 

\subsection{Hand-crafted features}
\label{subsec:features}
Apart from BERT embeddings, we also added 69 features which can be grouped into several categories:

\begin{itemize}
\item Neuralcoref,\footnote{\url{https://github.com/huggingface/neuralcoref}} Stanford CoreNLP \cite{CoreNLP} and e2e-coref \cite{e2e-coref} model predictions. It turned out that these models performed not really well in the task in hand, but their predictions worked well as additional features. 
\item Predictions of a Multi-Layered Perceptron trained with ELMo \cite{elmo} embeddings 
\item Syntactic roles of entities A, B, and Pronoun (subject, direct object, attribute etc.) extracted with SpaCy \footnote{\url{https://spacy.io/}}. 
\item Positional and frequency-based (distances between A, B, Pronoun and derivations, whether they all are in the same sentence or Pronoun is in the following one etc.). Many of these features we motivated by the Hobbs algorithm \cite{Hobbs1986} for coreference resolution. 
\item Named entities predicted for A and B with SpaCy
\item GAP heuristics outlined in the corresponding paper \cite{gap_paper} and briefly discussed in Sec. \ref{sec:gap_paper}
\end{itemize}

We need to mention that adding all these features had only minor effect on the quality of pronoun resolution (resulted in a 0.01 decrease in logarithmic loss when measured on the Kaggle test dataset) as compared to e.g. fine-tuning BERT classifier.

\subsection{Neural network architectures}
\label{sec:nns}

Final setup includes:

\begin{itemize}
\item 6 independently trained fine-tuned BERT classifiers with preprocessing described in Subsec. \ref{subsec:finetune}. In Tables \ref{table:logloss_kaggle}, \ref{table:gap_test_res}, and \ref{table:gap_test_res_corrected}, we refer to their averaged prediction as to that of a ``fine-tuned'' model (\fire)
\item 5 multi-layered perceptrons trained with different combinations of BERT embeddings for A, B, Pronoun (see Subsec. \ref{subsec:bert_preprocess}) and hand-crafted features (see Subsec. \ref{subsec:features}), all together referred to as ``frozen'' in Tables \ref{table:logloss_kaggle}, \ref{table:gap_test_res}, and \ref{table:gap_test_res_corrected} (\frozen ). Using MLPs with pre-trained BERT embeddings is motivated by \cite{tenney2019}. Two MLPs- separate for Cased and Uncased BERT models - both taking 9216-d input and outputting 112-d vectors. Two Siamese networks were trained on top of distances between Pronoun and A-embeddings, Pronoun and B-embeddings as inputs. One more MLP took only 69-dimensional feature vectors as an input. Finally, a single dense layer mapped outputs from the mentioned 5 models into 3 classes corresponding to named entities A, B or ``Neither''.
\item Blending (\blend) involves taking predicted probabilities for A, B and ``Neither'' with weight 0.65 for the ``fine-tuned'' model and summing the result with 0.35 times corresponding probabilities output by the ``frozen'' model.
\end{itemize}

In the next Section, we perform the analysis identical to the one done in \cite{gap_paper} to measure the quality of pronoun resolution and the severity of gender bias in the task in hand. 

\subsection{Correcting mislabeled instances}
\label{sec:corrections}
During the competition, 158 label corrections were proposed for the GAP dataset\footnote{\url{https://www.kaggle.com/c/gendered-pronoun-resolution/discussion/81331}} - when Pronoun is said to mention A but actually mentions B and vice versa. For the GAP test set, this resulted in 66 pronoun coreferences being corrected. It's important to mention that the observed mislabeling is a bit biased against female pronouns (39 mislabeled feminine pronouns versus 27 mislabeled masculine ones), and it turned out that most of the gender bias for F1 score and accuracy comes from these mislabeled examples.

\section{Results}
\label{sec:results}

In Table \ref{table:logloss_kaggle}, we report logarithmic loss that we got on GAP test (``gap-test.tsv''), and Kaggle test (Stage 2) datasets. Kaggle competition results can also be seen on the final competition leaderboard.\footnote{\url{https://www.kaggle.com/c/gendered-pronoun-resolution/leaderboard}} We report GAP test results as well to further compare with the results reported in the GAP paper: measured are logarithmic loss, F1 score and accuracy for masculine and feminine pronouns (Table \ref{table:gap_test_res}). Logarithmic loss and accuracy are computed for a 3-class classification problem (A, B, or Neither) while F1 is computed for a 2-class problem (A or B) to compare with results reported by the Google AI Language team in \citep{gap_paper}.

\begin{table}[t!]
\centering
\begin{tabular}{|l|c|c|}
\hline
                    & \textbf{GAP test} & \textbf{Kaggle test} \\ \hline
\fire \ \textbf{fine-tuned} & 0.29              & 0.192                \\ \hline
\frozen \ \textbf{frozen}     & 0.299             & 0.226                \\ \hline
\blend \ \textbf{blend}      & 0.257             & 0.185                \\ \hline
\end{tabular}
\caption{Logarithmic loss reported for the GAP test set, and Kaggle test (Stage 2) data for the model with fine-tuned BERT classifier (\fire), MLPs with pre-trained BERT embeddings and hand-crafted features (\frozen) and a blend of the previous two (\blend). There are 66 corrections done for GAP test labels as described in Subsec. \ref{sec:corrections}.}
\label{table:logloss_kaggle}
\end{table}

\begin{table*}[t!]
\centering
\scalebox{0.8}{
\begin{tabular}{|l|l|l|l|l|l|l|l|l|l|l|l|l|}
\hline
\multicolumn{1}{|c|}{\textbf{}} & \multicolumn{4}{c|}{\textbf{Logarithmic loss}}                                                                                        & \multicolumn{4}{c|}{\textbf{Accuracy}}                                                                                                & \multicolumn{4}{c|}{\textbf{F1 score}}                                                                                                \\ \hline
\multicolumn{1}{|c|}{}          & \multicolumn{1}{c|}{\textbf{M}} & \multicolumn{1}{c|}{\textbf{F}} & \multicolumn{1}{c|}{\textbf{O}} & \multicolumn{1}{c|}{\textbf{B}} & \multicolumn{1}{c|}{\textbf{M}} & \multicolumn{1}{c|}{\textbf{F}} & \multicolumn{1}{c|}{\textbf{O}} & \multicolumn{1}{c|}{\textbf{B}} & \multicolumn{1}{c|}{\textbf{M}} & \multicolumn{1}{c|}{\textbf{F}} & \multicolumn{1}{c|}{\textbf{O}} & \multicolumn{1}{c|}{\textbf{B}} \\ \hline
\fire \ \textbf{fine-tuned}             & 0.294                           & 0.398                           & 0.346                           & 0.738                           & 0.908                           & 0.884                           & 0.896                           & 0.974                           & 0.927                           & 0.9                             & 0.914                           & 0.971                           \\ \hline
\frozen \ \textbf{frozen}                 & 0.308                           & 0.368                           & 0.338                           & 0.837                           & 0.883                           & 0.866                           & 0.874                           & 0.981                           & 0.904                           & 0.882                           & 0.893                           & 0.976                           \\ \hline
\blend \ \textbf{blend}                  & 0.259                           & 0.338                           & 0.299                           & 0.766                           & 0.907                           & 0.883                           & 0.895                           & 0.974                           & 0.923                           & 0.898                           & 0.911                           & 0.973                           \\ \hline
\end{tabular}
}
\caption{Performance of the proposed two models and their blending on the GAP test set, split by \textbf{M}asculine, \textbf{F}eminine (\textbf{B}ias shows F/M in case of F1 and accuracy, and M/F in case of logarithmic loss), and \textbf{Overall}.}
\label{table:gap_test_res}
\end{table*}

We also incorporated 66 label corrections as described in \ref{sec:corrections} and, interestingly enough, this lead to a conclusion that with corrected labels, models are less susceptible to gender bias. Table \ref{table:gap_test_res_corrected} reports the same metric in case of corrected labeling, and we see that in this case the proposed models are almost gender-unbiased. 

\begin{table*}[t!]
\centering
\scalebox{0.8}{
\begin{tabular}{|l|l|l|l|l|l|l|l|l|l|l|l|l|}
\hline
\multicolumn{1}{|c|}{} & \multicolumn{4}{c|}{\textbf{Logarithmic loss}}                                                                                        & \multicolumn{4}{c|}{\textbf{Accuracy}}                                                                                                & \multicolumn{4}{c|}{\textbf{F1 score}}                                                                                                \\ \hline
\multicolumn{1}{|c|}{} & \multicolumn{1}{c|}{\textbf{M}} & \multicolumn{1}{c|}{\textbf{F}} & \multicolumn{1}{c|}{\textbf{O}} & \multicolumn{1}{c|}{\textbf{B}} & \multicolumn{1}{c|}{\textbf{M}} & \multicolumn{1}{c|}{\textbf{F}} & \multicolumn{1}{c|}{\textbf{O}} & \multicolumn{1}{c|}{\textbf{B}} & \multicolumn{1}{c|}{\textbf{M}} & \multicolumn{1}{c|}{\textbf{F}} & \multicolumn{1}{c|}{\textbf{O}} & \multicolumn{1}{c|}{\textbf{B}} \\ \hline
\fire \ \textbf{fine-tuned}    & 0.268                           & 0.311                           & 0.29                            & 0.863                           & 0.914                           & 0.905                           & 0.91                            & 0.99                            & 0.932                           & 0.919                           & 0.926                           & 0.987                           \\ \hline
\frozen \ \textbf{frozen}        & 0.292                           & 0.306                           & 0.299                           & 0.954                           & 0.886                           & 0.89                            & 0.888                           & 1.005                           & 0.908                           & 0.906                           & 0.907                           & 0.997                           \\ \hline
\blend \ \textbf{blend}         & 0.241                           & 0.273                           & 0.257                           & 0.882                           & 0.913                           & 0.908                           & 0.91                            & 0.995                           & 0.928                           & 0.921                           & 0.924                           & 0.992                           \\ \hline
\end{tabular}
}
\caption{Performance of the proposed two models and their blending on the GAP test set with 66 corrected labels, split by \textbf{M}asculine, \textbf{F}eminine (\textbf{B}ias shows F/M in case of F1 and accuracy, and M/F in case of logarithmic loss), and \textbf{Overall}.}
\label{table:gap_test_res_corrected}
\end{table*}

These results imply that:
\begin{itemize}
\item Overall, in terms of F1 score, the proposed solution compares very favorably with the results reported in the GAP paper, achieving as high as $0.911$ overall F1 score, compared to $0.729$ for ``Parallelism + URL'' heuristic from \citep{gap_paper}; 
\item Blending model predictions improves logarithmic loss pretty well but does not impact F1 score and accuracy that much. It can be explained: logarithmic loss is high for confident and at the same time incorrect predictions. Blending averages predicted probabilities so that they end up less extreme (not so close to 0 or 1);
\item With original labeling, all models are somewhat susceptible to gender bias, especially in terms of logarithmic loss. However, in terms of F1 score, gender bias is still less than for e2e-coref and ``Parallelism + URL'' heuristic reported in \cite{gap_paper};
\item Fixing some incorrect labels almost eliminates gender bias, when we talk about F1 score and accuracy of pronoun resolution. 
\end{itemize}

\section{Conclusions and further work}
\label{sec:future}
We conclude that we managed to propose a BERT-based approach to pronoun resolution which results in considerably better quality (as measured in terms of F1 score and accuracy) than in case of pronoun resolution done with heuristics described in the GAP paper. Moreover, the proposed solution is almost gender-unbiased - pronoun resolution is done almost equally well for masculine and feminine pronouns. 

Further we plan to investigate which semantic and syntactic information is carried by different BERT layers and how it refers to coreference resolution. We are also going to benchmark our system on OntoNotes, Winograd, and DPR datasets.

\section*{Acknowledgments}

Authors would like to thank Open Data Science\footnote{\url{https://ods.ai}} community for all insightful discussions related to Natural Language Processing and, more generally, to Deep Learning. Authors are also grateful to Kaggle and Google AI Language teams for organizing the Gendered Pronoun Resolution challenge. 

\bibliography{bert_unbiased_pronoun_resolution}
\bibliographystyle{acl_natbib}

\end{document}